# CONTRAST ENHANCEMENT AND BRIGHTNESS PRESERVATION USING MULTI-DECOMPOSITION HISTOGRAM EQUALIZATION

Sayali Nimkar, Sucheta Shrivastava and Sanal Varghese

Department of Electronics and Telecommunication Engineering,
Atharva college of Engineering, Maharashtra, India.
nimkar.sayali@gmail.com
suchetashrivastava@yahoo.co.in
sanalalice@gmail.com

## ABSTRACT

*Histogram Equalization (HE) has been an essential addition to the Image Enhancement world. Enhancement techniques like Classical Histogram Equalization(CHE),Adaptive Histogram Equalization (AHE), Bi-Histogram Equalization (BHE) and Recursive Mean Separate Histogram Equalization (RMSHE) methods enhance contrast, brightness is not well preserved, which gives an unpleasant look to the final image obtained. Thus, we introduce a novel technique Multi-Decomposition Histogram Equalization (MDHE) to eliminate the drawbacks of the earlier methods. In MDHE, we have decomposed the input image using a unique logic, applied CHE in each of the sub-images and then finally interpolated them in correct order. The final image after MDHE gives us the best results based on contrast enhancement and brightness preservation aspect compared to all other techniques mentioned above. We have calculated the various parameters like PSNR, SNR, RMSE, MSE, etc. for every technique. Our results are well supported by bar graphs, histograms and the parameter calculations at the end.*

## 1. INTRODUCTION.

Image processing is avast and challenging domain with its applications in fields like medical, aerial and satellite images, industrial applications, law enforcement, and science. Often the quality of an image is more often linked to its contrast and brightness levels enhancing these parameters will certainly give us the best result. Our main area of research is MDHE for Histogram Equalization (HE).Here, HE is an image enhancement method that allocates the pixel values evenly, thus developing a better picture. Image Enhancement majorly involves four key parameters – [1] brightness –Brightness can be modified by increasing „gamma". Gamma is a non-linear form of increase in brightness. [2] contrast- It is the separation between the dark and bright areas of an image. Thus, increasing contrast increases darkness in dark areas and brightness in bright areas. [3] Saturation- Saturation is increasing the separation between the shadows and highlights. [4] Sharpness– It is related to edges, the contrast along the edges of a photo. Using histogram equalization contrast can be enhanced. It is a straightforward and Invertible operator. There are various histogram equalization techniques with their own advantages and disadvantages. Our method Multide composition histogram equalization however is a unique combination of CHE and types of Histogram Equalization.

## 2. HISTOGRAM EQUALIZATION TECHNIQUES

There are numerous methods by which Histogram of an image can be equalized. Depending upon the area of Application, we can choose the different histogram equalization techniques. We will see the following four types of Histogram Equalization methods in detail:

2.1 Classical Histogram Equalization (CHE)

2.2 Adaptive Histogram Equalization (AHE)

2.3 Bi- Histogram Equalization (BHE)

2.4 Recursive Mean Separate Histogram Equalization (RMSHE)

2.5 Multi-Decomposition Histogram Equalization (MDHE)

### 2.1 Classical Histogram Equalization

CHE is the fundamental technique for image processing, especially when gray level images are considered. The aim of this method is to distribute the given number of gray levels over a range uniformly, thus enhancing its contrast. The cumulative density function (CDF) is formulated by the below mentioned expression:

$$C_i^{O[l_s, l_f]} = \sum_{i=l_s}^{l_f} P_i^{I[l_s, l_f]} = \frac{i - l_s + 1}{(l_f - l_s + 1)}.$$

The CHE tries to produce an output image with a flattened histogram, means a uniform distribution. An image is formed by the dynamic range of values of gray levels. Basically, the entire gray levels are denoted as 0 to $L - 1$.

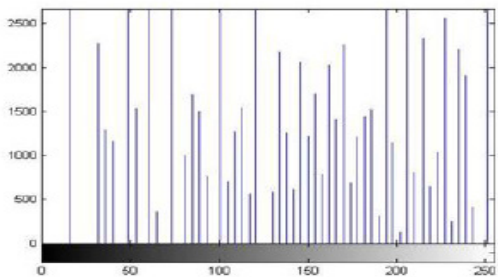

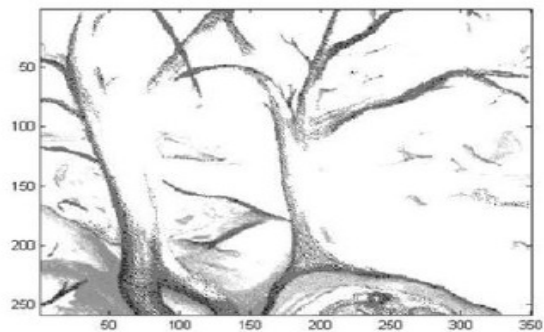

Figure 1.Histogram after CHE                    Figure 2. Image after CHE

**2.1.1 Disadvantage**

1. A disadvantage of this method is that it is undifferentiating between the various pixels, that is, while increasing the contrast of its background, the signal gets distorted.

2. Histogram equalization often produces unrealistic and unlikely effects in photographs.

## 2.2 Adaptive Histogram Equalization

Adaptive Histogram Equalization (AHE) is used to improve contrast in images. It computes many ordinary histograms, each one analogous with a section of the image. Thus, the output results in each to redistributing the lightness values. It is appropriate to adjust the local contrast and to fetch clear details.

On the other hand, AHE is responsible for over-amplifying noise in some homogeneous regions of an image. To avoid this drawback, an advanced version of AHE, called Contrast Limited Adaptive Histogram Equalization (CLAHE) is introduced.

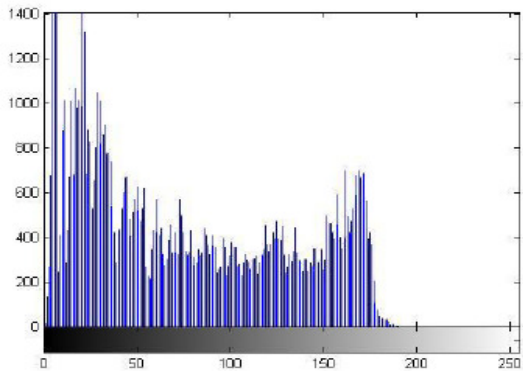   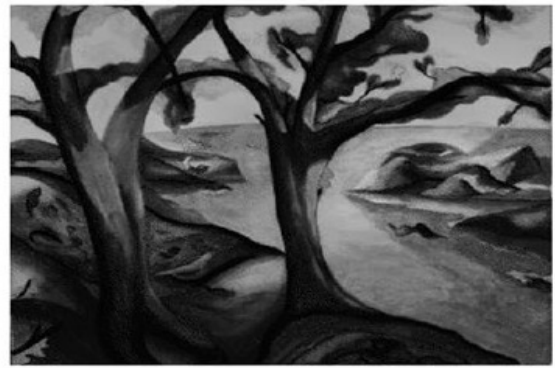

Figure 3.Histogram after AHE          Figure 4.Image after AHE

**2.2.1 Disadvantage**

- AHE has a behavior of amplifying noise, thus limiting its use for homogeneous figures.
- Its advanced form is contrast limited adaptive histogram equalization (CLAHE) that eliminated the above problem.
- It also fails to retain the brightness with respect to the input image.

## 2.3 Bi-Histogram Equalization

The major basis of origination of this method is to overcome the drawback introduced by CHE. Here, the original image is segmented twice i.e. into two sub-sections. This is done by dividing the meangray level and then applying CHE method on each of the two sub-sectioned image. Its objective is to produce method suitable for real-time applications. But again this method has the same disadvantage as CHE by inputting unwanted signals.

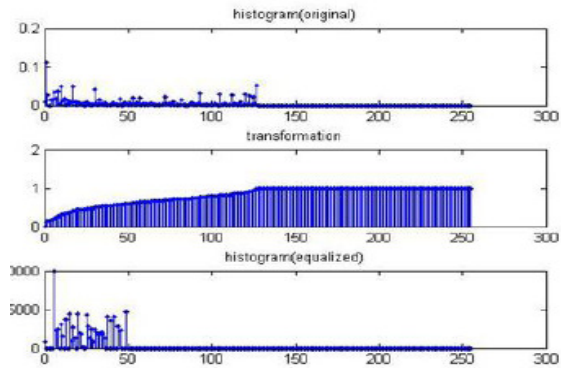
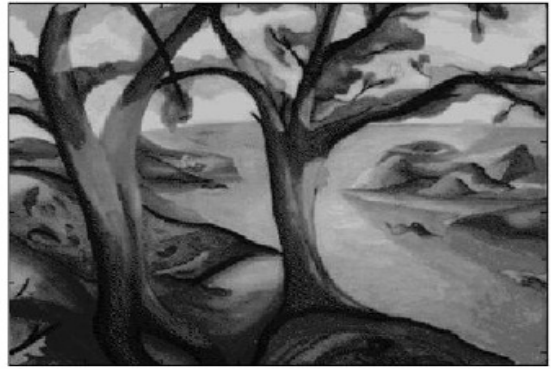

Figure 5. Histogram for BHE                 Figure 6.Image after BHE

## 2.4 Recursive Mean Separate Histogram Decomposition

An extended version of the BHE method proposed before, and named as recursive mean-separate HE(RMSHE), proposes the following. Instead of decomposing the image only once, the RMSHE method offers to perform image decomposition recursively, up to a scale r, generating 2r sub-images. After, each one of these sub-images is independently enhanced using the CHE method. Note that, computationally speaking, this method presents a problem: the number of decomposed sub-histograms is a power of two.

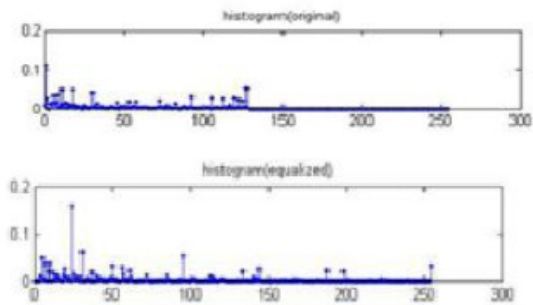
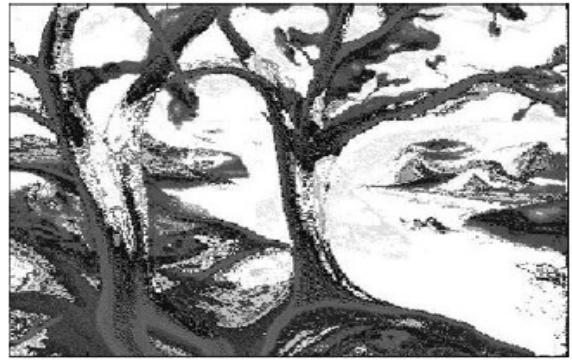

Figure 7.Histogram of RMSHE                 Figure 8.Image of RMSHE

## 2.5 Multi-Decomposition Histogram Equalization

All the HE methods that we have covered prior to this, enhances the contrast of an image but are unable to preserve its brightness. As a result, these methods can generate unnatural and non-existing objects in the processed image. To eliminate these limitations, MDHE comes up with a novel technique by decomposing the image into various small images. Then the image contrast enhancement provided by CHE in each sub-image is less concentrated, leading the output image to have a more likely and acceptable look.

We have followed a four step process to carry out our technique effectively:

### 2.5.1 Multi-Decomposition of the image

An image is taken as input and divided into as many as 64 sub-images (it is flexible according to application field). This is implemented using the spatial domain techniques. Functions are called and the decomposition of the image is done.

### 2.5.2 Applying histogram based techniques

Now after dividing the image into 64 sub-images we apply Adaptive Histogram Equalization method on each of the 64 sub-images to obtain enhanced sub-images. This is implemented by using nested for loop.

### 2.5.3 Interpolating the image

The next part that is to be done is to interpolate all the sub-images in the right sequence, carefully at the right place to get our Interpolated image. Though contrast enhancement has been achieved, the image still lacks brightness preservation.

### 2.5.4 Brightness Preservation

To preserve the brightness we now apply a code according to which we can set a limit which preserve brightness. Therefore, at the end of the entire process, we have obtained an image which is contrast enhanced, brightness preserved as well as there is a natural look to the image. This distinguishes our method from the others.

Thus the output images obtained by applying MDHE give amazing results, thus satisfying our need to select this method. The images, bar graphs and histograms of the entire project are attached in the conclusion. This will help the reader to understand the results in a much better and effective way.

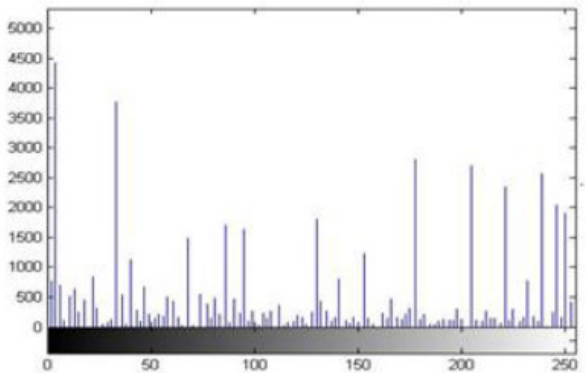
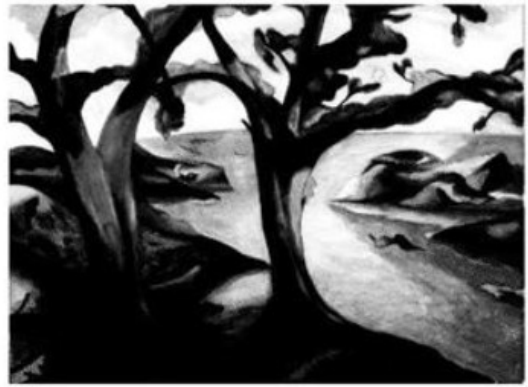

Figure 9.Equalized Histogram of MDHE        Figure 10. Image of MDHE

## 2.5.5 MDHE Flowchart:

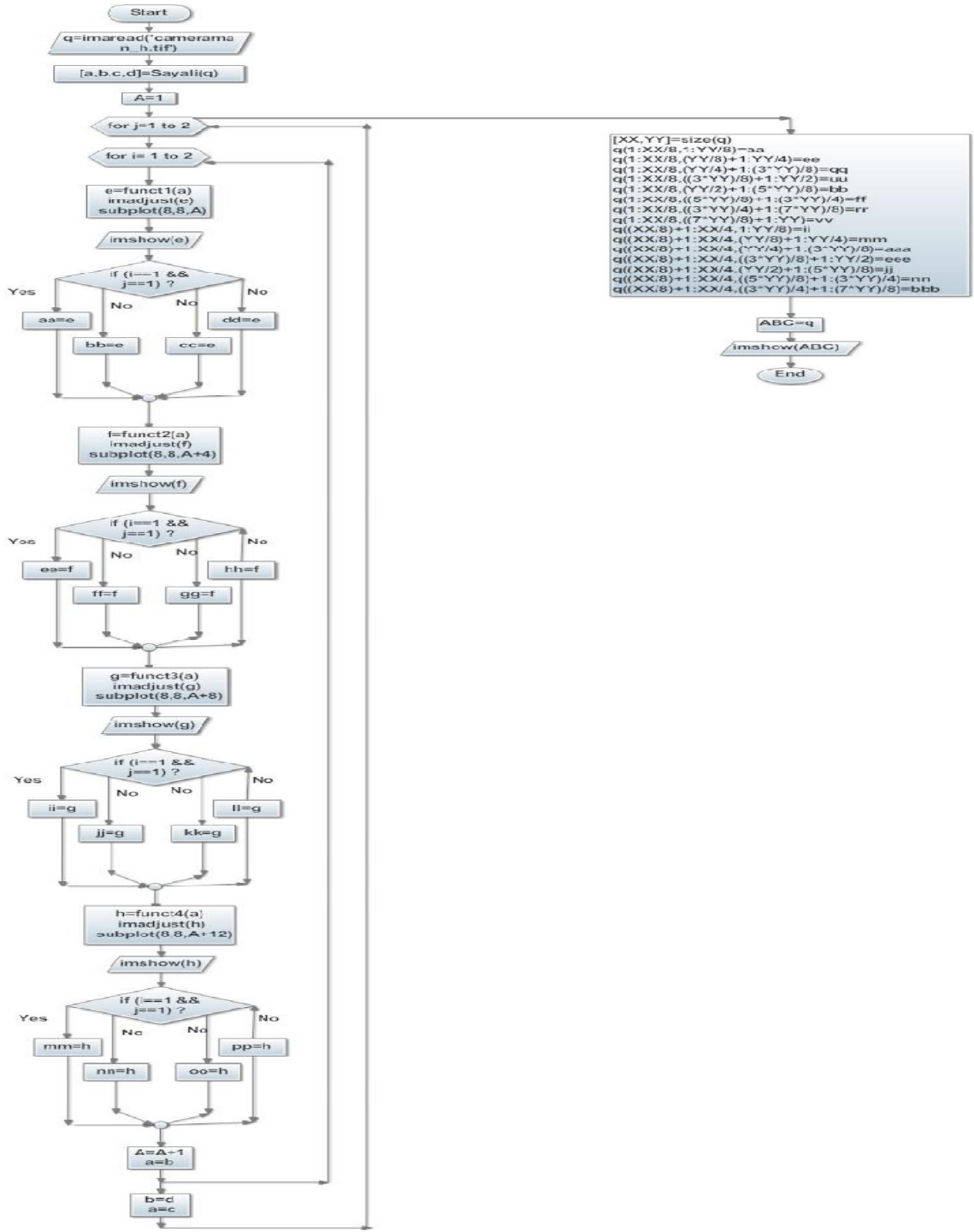

Figure11. MDHE flowchart

# 3. DISCUSSION

The performance of Multi-Decomposition Histogram Equalization (MDHE) is measured using Image Enhancement Parameters such as Mean Absolute error (MAE), Pearson correlation Coefficient (PCC), Signal to Noise Ratio (SNR), Peak Signal to Noise Ratio (PSNR), Mean Squared Error (MSE) and Root Mean squared error (RMSE). They help in evaluating the effectiveness of the Image enhancement technique thus used.

## 3.1 Image Enhancement Parameters

### 3.1.1 Signal to Noise Ratio (SNR) is

It gives us the relation betweenrequired signal levelandsurrounding noise level. It is defined as the ratio of signal power to noise power. A ratio of higher than 1:1 is regarded as a well signaled ratio. It is measured in Decibels. Represented as:

$$\text{SNR}_{\text{dB}} = 10 \log_{10} \left( \frac{P_{\text{signal}}}{P_{\text{noise}}} \right) = P_{\text{signal,dB}} - P_{\text{noise,dB}},$$

### 3.1.2 Peak Signal to Noise Ratio (PSNR) is

It is the fraction of the optimum power level to a desired signal and the optimized power of disturbance  noise that affects the reliability of its representation expressed in logarithmic decibel scale. It is generally used in measuring the quality of reconstruction done onlossy compression codecs.

### 3.1.3 Mean Squared Error (MSE) is

It deals with the values obtained by an estimator thus calculating the divergence between estimator values and optimum values of estimated quantity. MSE quantifies the average of squares of the "errors" .The higher value of MSE the better.

### 3.1.4 Root Mean Square Error (RMSE) is

It calculates the root of power two for Standard Deviation. It measures the average magnitude of the error. It is most useful when large errors are specifically undesirable. Given by:

$$RMSErrors = \sqrt{\frac{\sum_{i=1}^{n} (\hat{y_i} - y_i)^2}{n}}$$

### 3.1.5 Pearson Correlation Coefficient(PPMCC or PCC) is

In statistics, the Pearson product-moment correlation co-efficient is denoted by r and it measures the correlation i.e. the strength of linear dependence between two variables X and Y, giving a value between +1 and -1 inclusive.

### 3.1.6 Universal Quality index for images (UIQ) is

It is calculated by structuring any image abnormality as an amalgamation of parameters such as correlation loss, luminance distortion and contrast distortion. It performs considerably better than the widely used distortion metric mean squared error. And it exhibits consistency with subjective quality measurement on various models and experiments employed.

### 3.1.7 Mean Absolute Error (MAE) is

It is used to measure how close forecasts or our predictions are to the eventual outcomes. Mean Absolute error is given by:

$$\text{MAE} = \frac{1}{n}\sum_{i=1}^{n}|f_i - y_i| = \frac{1}{n}\sum_{i=1}^{n}|e_i|$$

It amounts for the accuracy for continuous variables.

## 4. RESULT

The improvement in the quality and clarity in image is clearly seen in the figure of the Tree as can be made out from above examples. Clearly, The Tree in MDHE looks well contrasted and brightened. Another comparison of "Sunset.jpg" with Table- I values highest in Pearson Co-efficient and PSNR value for MDHE that give it the edge over others. For MSE and RMSHE, the values of CHE are 125.2 and 11.89 respectively nearing to it is MDHE. With the other MAE, SNR and UIQ values averaging to the finest. Rounding off to precision is the MDHE amongst the other methods.

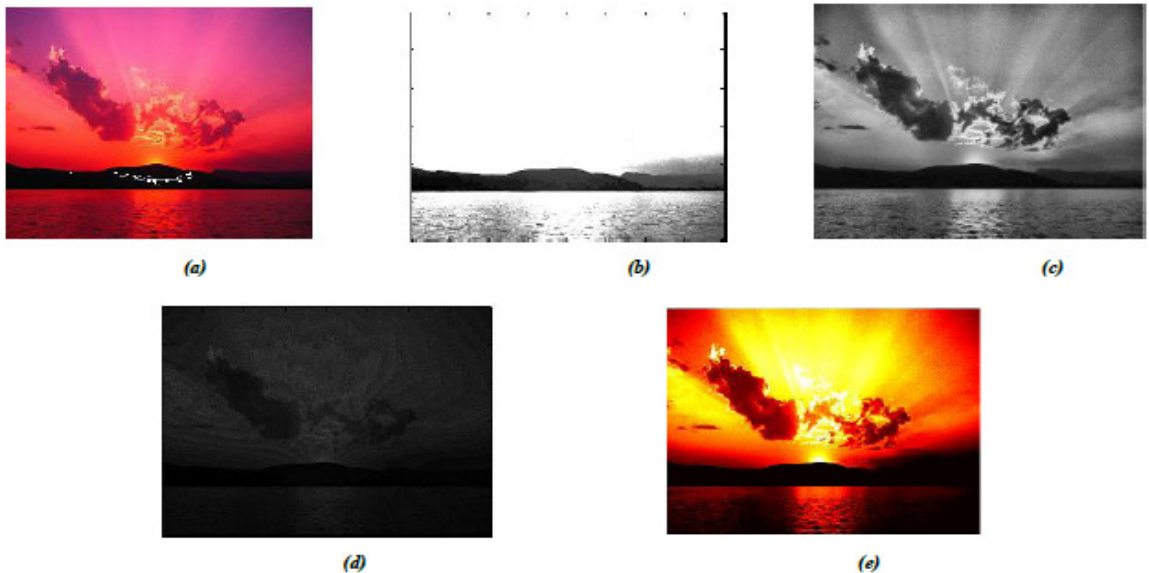

Figure 12. Comparison of "Sunset.jpg" using various methods(a) CHE (b) AHE (c) BHE
(d) RMSHE (e) MDHE

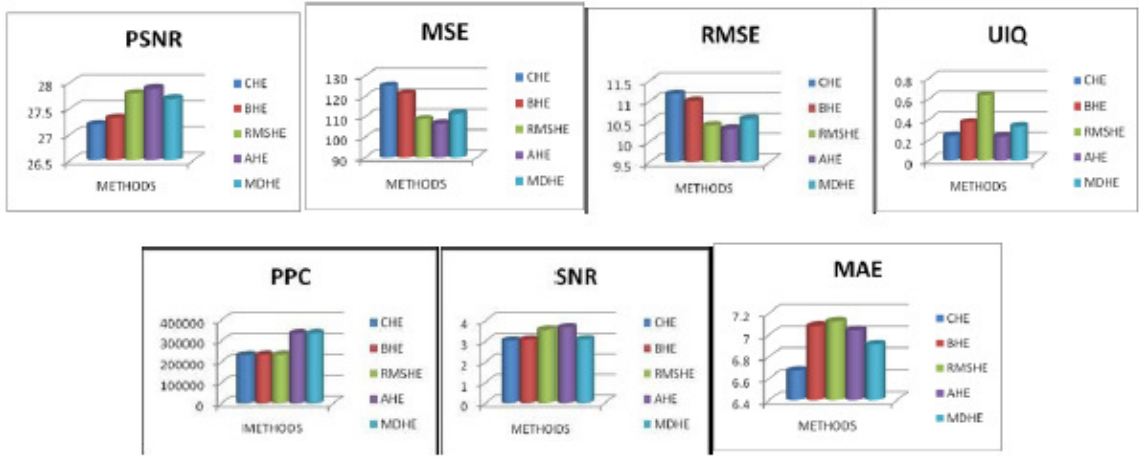

Figure 13. Graphical representation for the image "Sunset.jpg" using various image enhancement parameters.

TABLE 1: Comparison of parameters for "Sunset.jpg" using various methods

| Parameter/method | CHE | BHE | RMSHE | AHE | MDHE |
|---|---|---|---|---|---|
| PSNR | 27.18847 | 27.32256 | 27.79624 | 27.88401 | **27.68765** |
| MSE | **125.20817** | 121.40150 | 108.85680 | 106.67901 | **111.61293** |
| RMSE | **11.18965** | 11.01823 | 10.43345 | 10.32855 | **10.56470** |
| UIQ | 0.24200 | 0.37324 | **0.63568** | 0.23664 | **0.33402** |
| PPC | 232005.43 | 233379.44 | 233608.79 | 336458.40 | **336898.90** |
| SNR | 3.01523 | 3.06608 | 3.55238 | 3.67469 | 3.07708 |
| MAE | 6.67066 | 7.08924 | 7.12072 | 7.04791 | 6.91421 |

## 5. CONCLUSIONS AND FUTURE APPLICATIONS

HE works on the four main elements of images: saturation, contrast, sharpness and brightness. We focus on these four parameters and thus, enhance the quality of images. We obtain the desired contrast levels, along with preservation of brightness and not only this, the natural look of the input image is maintained. Future applications include photos obtained from satellite communications – since we obtain images from satellite that are distorted due to space interference and dispersion losses. Other application fields are Medical field- X-Rays, Meteor descriptions, Discoveries of Geo-stationary bodies, weather information, law enforcement that includes finger processing, surveillance camera processing. Science- enhancing electron microscope image for readability.

**AUTHORS**

Sanal Varghese : Completed bachelors at Atharva College Of Engineering. Working at Infosys Ltd, Mysore. Areas of research comprise of Image processing, Wireless networks and Data communication.

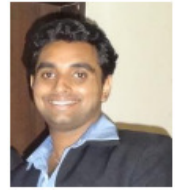

Sayali Nimkar : Pursuing MS in Electrical Engineering at University of North Carolina, Charlotte.Interests include Signal processing and Wireless sensor networks.

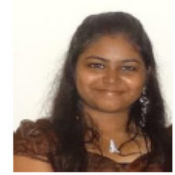

Sucheta Shrivastava : Student at Atharva college of engineering. Pursuing masters. Interests include Signal & Image processing, semiconductors, wired and wireless networking and optics .

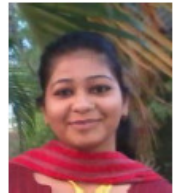